\documentclass{article}

\usepackage[final]{nips_2018}

\usepackage[utf8]{inputenc} 
\usepackage[T1]{fontenc}    
\usepackage{hyperref}       
\usepackage{url}            
\usepackage{booktabs}       
\usepackage{amsfonts}       
\usepackage{nicefrac}       
\usepackage{microtype}      
\usepackage{array}
\usepackage{slashbox}

\relax
\usepackage{times}  
\usepackage{helvet}  
\usepackage{courier}  

\usepackage{graphicx}  
\usepackage{mathrsfs}
\usepackage{times,algpseudocode}
\usepackage{latexsym}
\usepackage{url}
\usepackage{slashbox}
\usepackage{amssymb}
\usepackage{amsmath}
\usepackage{latexsym}
\usepackage{tablefootnote}
\usepackage{pdflscape}
\usepackage{booktabs}
\usepackage{soul}
\usepackage{xcolor,colortbl}
\usepackage{amsfonts}
\usepackage[english]{babel}
\usepackage{tikz-dependency}
\usepackage{tikz}
\usepackage{tikz-qtree}
\usepackage{tablefootnote}
\usepackage{multibib}
\usepackage{algorithm}
\usepackage{wrapfig,lipsum,booktabs,multirow,subcaption,multicol}
\usepackage{xcolor,colortbl,framed,cuted,tcolorbox}
\usepackage{pbox}

\title{A New NMT Model for Translating Clinical Texts from English to Spanish}

\author{
Rumeng Li$^{1}$,
Xun Wang$^{1}$,
Hong Yu$^{2,1,3,4}$ \\
\\
$^{1}$ College of Information and Computer Science, University of Massachusetts Amherst, Amherst, MA, USA \\
$^{2}$ Department of Computer Sciences, University of Massachusetts Lowell, Lowell, MA, USA \\
$^{3}$ Department of Quantitative Health Sciences, University of Massachusetts Medical School, Worcester, MA, USA \\
$^{4}$ Bedford VAMC and CHOIR, Bedford, MA, USA \\
\\
\texttt{rli@cs.umass.edu, xunwang@cs.umass.edu, hong\_yu@uml.edu}
}

\begin{document}


\maketitle

\begin{abstract}
Translating electronic health record (EHR) narratives from English to Spanish is a clinically important yet challenging task due to the lack of a parallel-aligned corpus and the abundant unknown words contained. To address such challenges, we propose \textbf{NOOV} (for No OOV), a new neural machine translation (NMT) system that requires little in-domain parallel-aligned corpus for training. NOOV integrates a bilingual lexicon automatically learned from parallel-aligned corpora and a phrase look-up table extracted from a large biomedical knowledge resource, to alleviate both the unknown word problem and the word-repeat challenge in NMT, enhancing better phrase generation of NMT systems. Evaluation shows that NOOV is able to generate better translation of EHR with improvement in both accuracy and fluency. 
\end{abstract}

\section{Introduction}
Providing language-appropriate access to EHR narratives is important for patients, as better understanding of their medical conditions could promote communication between physicians and patients and improve patients' medication adherence and self-managed care. In the Unites States, most EHRs are written in English. However, there are over 37.6 million people who speak Spanish at home, of which 16.5 million report speaking English less than very well \cite{ryan2013language}. Hospitals face challenges in providing translated EHR narratives due to the time, work burden, expenses of translating service and privacy concerns, thus the development of an efficient and secure machine translation system is needed for the US hospitals. In contrast to the active research of MT in the open-domain, relatively few MT systems have been developed in the biomedical domain with early work mostly on SMT systems \cite{eck2004improving,liu2006babelmesh,merabti2011translating,zeng2010can,liu2015translating,bojar2014findings} and later on NMT methods \cite{wolk2015neural,yepes2017findings}.

Neural machine translation (NMT) methods have recently achieved promising results 
\cite{kalchbrenner2013recurrent,cho2014learning,bahdanau2014neural,vaswani2017attention,gehring2017convolutional}.
A major drawback of NMT systems is their inability to correctly translate rare, out-of-vocabulary (OOV) or unknown words. Most NMT systems limit the vocabulary to frequent words and convert rare words into a single \textit{unk} (unknown words) symbol to reduce computational complexity. Such NMT models fail to learn the translation of unknown words. Many existing work also reported that sentences with many rare words tend to be translated poorly \cite{sutskever2014sequence,bahdanau2014neural,li2016towards}. Various approaches have been proposed to address the unknown word problem in NMT including building representations for unknown words based on characters and other sub-word units \cite{belinkov2017synthetic,ling2015character,luong2016achieving,lee2016fully,zhao2016deep,sennrich2015neural}.  
Lexicon/dictionary information is also used in existing work \cite{arthur2016incorporating,luong2014addressing}. Another approach aims to replace unknown words by copying relevance source words through alignment tools \cite{luong2014addressing} or attention scores \cite{hashimoto2016domain,jean2015montreal}, or replace the rare words with similar in-vocabulary words \cite{li2016towards}. The rare word problem is even more serious for the clinical domain. The clinical text is a sub-language, and it has a rich domain-specific vocabulary including complex disease and drug names, medical terms, shortened forms, and other domain-specific jargon. For example, in our EHR corpus, unknown words are as high as 25.6\%. Furthermore, in part due to the high-cost and HIPAA regulation, there is no parallel-aligned EHR corpus available, making it harder to develop an NMT system. Existing parallel corpora for biomedical machine translation are generally literature data automatically extracted from medical articles \cite{liu2015translating,yepes2017findings}, which are quite different from the EHR data.  

To solve the above problems, we first created a benchmark parallel-aligned English-Spanish EHR corpus. Using this corpus as the gold standard, we evaluate a new NMT system that integrates a bilingual lexicon automatically learned from open-domain parallel corpora and a phrase look-up table extracted from a biomedical knowledge resource. Unknown words due to its infrequency makes the NMT system difficult to learn sufficient information for proper translation. The lexicon, on the other hand, can remedy this problem by assigning probability to every aligned occurrence. The incorporation of the lexicon serves as an extra guide for the decoder. The incorporation of the phrase look-up table comes in the process of decoding, instead of usual ways of post-processing. In addition to improving phrase translation, the incorporation of the phrase look-up table helps guide the decoder in the following step, without damaging the decoder's language model, hence reduce the word repetition problem in NMT \cite{feng2016implicit,lin2018decoding}. We further conduct experiments to verify the effectiveness of our NOOV model in addressing the unknown words problem. 

\section{Methods}
The proposed model adopts the encoder-decoder scheme and uses attention mechanism. 
\paragraph{Encoding}

Same with the seq2seq model, we use a biLSTM to scan the input $s=(w^s_1,w^s_2,...,w^s_m)$ to generate its vector representations:
\begin{equation}
o^f_i, h^f_i = \text{biLSTM}(w_i, h^f_{i-1}); \;
o^b_i, h^b_i = \text{biLSTM}(w_i, h^b_{i-1}); \;
h_i = [h^f_i : h^b_i]
\end{equation}

\paragraph{Lexicon and phrase look-up table assisted decoding}

Then the decoder generates the target sentence word by word based on the encoded vector. Attention mechanism enables the decoder to look at the input again and chooses the most relevant part to attend to at each step in translation. 
\begin{equation}
\begin{aligned}
u'_{i}&=\sum_j att_j^i*h_j; att_{j}^i =g(h_{*},h'_{i-1},w'_{i-1})\\
prob^i(tok_k^t)&=\frac{exp^{pr^i(tok_k^t)}}{\sum_k(exp^{pr^i(tok_k^t)})}; pr^i(tok_k^t)=\sum_{tok_j^s  \in V^s}(att^i_j*p_k^j)\\
w'_i&=\arg\max ((1-\alpha)*v(o'_i)+\alpha*prob^i); o'_i,h'_i=LSTM(w'_{i-1},u'_{i})\\
\end{aligned}
\end{equation}

During decoding, at the $i^{th}$ step, the hidden state $h'_i$ of the decoder and the hidden states of the encoder $\{h_0, h_1, ..., h_i, ..., h_m\}$ are leveraged to calculate the attention weights.
$att^i_j$ is the weight assigned to $h_j$ in the $i^{th}$ step. It is also regarded as the probability of word $w_j^s$ in the source sentence aligned to the word $w_i^t$ in the target sentence. We leverage these $att_*^i$ to encode lexicon information for better translation.

The lexicon $\mathcal{L}$ we built contains alignments between source words  and target words. $lex(tok_j^s)=\{tok_k^t:p_k^j,tok_l^t:p_l^j, ...\}$. 
$lex(tok_j^s)$ is a set of candidate target words associated with probabilities. $p_k^j$ is the probability of $tok_j^s$ in the source language mapping to $tok^t_k$ in the target language. One word in the source language can be aligned to several words in the target language. We collect $lex(w_j^s)$  for each word $w_j^s$ in the source sentence. The details of collecting $lex(w_j^s)$ are explained below. At the $i_{th}$ step, a probability $prob^i$ over all words in the target language is built using $lex(w_*^s)$ and $a^i_*$.

$V^s=\{tok_k^s\}$  and $V^t=\{tok_k^t\}$ are the source vocabulary and the target vocabulary. For each word in the target vocabulary, its probability is $prob^i(tok_k^t)$.

$p^j_k$ is the probability that $w_j^s$ is mapped to $tok_k^t$. Meanwhile, the decoder outputs a probability over the target vocabulary, which is written as $v(o'_i)$. The final output at the $i^{th}$ step is decided according to  $output_i=\alpha*prob^i(V^t)+(1-\alpha)v(o'_i)$. $\alpha$ is a parameter which balances the impact of lexicon information and that of the decoder's outputs. This parameter is tuned in the following experiments.

We enhance the output from the seq-to-seq model by the use of $prob(V^t)$. As a weighted sum of mapping between source words and target words,  $prob(V^t)$ is more robust in dealing with rare words. The infrequency of unknown words makes seq-to-seq model difficult to learn its corresponding language model at decoding and correct translation generation, the $prob(V^t)$ acts as an extra supervision to remedy such insufficiency.

At decoding, the attention-based seq-to-seq model sometimes tends to repeatedly output the same word when the attention mechanism consecutively focuses on the same position in the source sentence. Such repetition problem has also been studied in the NMT literature \cite{feng2016implicit,lin2018decoding}. In our work, the output at the $i^{th}$ step is decided according to the output of decoder and the $prob^i$ we obtain using $att_*^i$. If $\arg\max_j(att_j^i)= \arg\max_j(att_j^{i+1})$ which means two consecutive words in the target sentence $w_i^t,w_{i+1}^t$ corresponds to the same word in the source sentence, we will have $\arg\max(prob^i)=\arg\max(prob^{i+1})$, which further leads to 
$\arg\max(output_i)=\arg\max(output_{i+1})$. To avoid this problem, we use a phrase look-up table $T=\{ph_i^s:ph_i^t\}$ to help select the correct phrase. When the model detects that $\arg\max(output_i)=\arg\max(output_{i+1})$, we will use the word in the source sentence $w^s_*$ and the word predicted at the $i^{th}$ step, $w^t_i$, to search the phrase look-up table to find the most suitable candidate. We select the phrase $ph^t_i$ which contains $w^t_i$ and whose corresponding source phrase $ph^s_i$is contained in $sent^s$. If multiple $(ph^s_i:ph^t_i)$ exist, we select the one with the longest $ph^s_i$. The corresponding word in $ph^t_i$ is selected as the output. This strategy enables the system to not only correct repetitions but also pass suitable inputs to the decoder at the following step.

\paragraph{Context-aware lexicon probability construction and phrase look-up table extraction}
Our lexicon is built based on the SMT tool Giza++ \cite{och2003systematic} which aims to maximize the alignments and lexical translation probabilities $prob(src|tgt)$ by using the expectation maximization (EM) algorithm. The lexicon built in this way could reflect the global information contained in a corpus. We also tried ways to construct our lexicon ``locally". 
At decoding, when translating a source sentence $sent^s_k$, we collect all the sentence pairs $sent^s_j:sent^t_j$  whose $sent_j^s$ shares at least one word with  $sent_k^s$ . Then we extract $lex(w_*^s)$ from these sentence pairs following the method of \cite{och2003systematic}. The extracted $lex(w_*^s)$ reflects the context. One word $w^s_i$ in the source sentence $sent^s_k$ may appear in different context and aligns to different words. Here we only consider  $w_i^s$ which appears in context that is related to the current $sent_s^k$. This constraint considers the context and helps build context-aware lexicon. 

In addition, we also made use of a biomedical knowledge resources the Unified Medical Language System (UMLS) \cite{bodenreider2004unified}. We use the 2017AB version of the UMLS to extract additional English-Spanish lexicons/phrasebooks. We extract the \textbf{preferred terms} English and Spanish concepts which are assigned the same unique identifier (CUI). The resulting Spanish-English phrase look-up table contains 465,256 pairs of words and phrases. 

\section{Experimental setup}
\paragraph{Parallel-aligned biomedical and EHR corpora}
The first corpus used is the English-Spanish Parallel Aligned Biomedical Corpora as the $ESPAC_{MedlinePlus}$ corpus built in \cite{liu2015translating}. 
We also built an English-Spanish Parallel-Aligned EHR notes corpus, which comprises of 3,020 paired sentences from 57 de-identified EHR discharge summaries, randomly selected from patients with type 2 diabetes in a hospital. Translation was done by a professional medical translator whose first-language was Spanish, who has spent over 1000 hours to build the corpus, including back translation, a very costly task. We held 20\% of the data as a testing set, of the rest, 90\% were used for training and 10\% were used for development. Table 1 shows the statistics of the two datasets.

\begin{table*}[t!]
\begin{center}
\small
\caption{\label{medline_data} Statistics of $ESPAC_{MedlinePlus}$ and $EHR$ datasets.}
\begin{tabular}{|l|c|c|c|c|c|}
\hline 
$ESPAC_{MedlinePlus}$ & Sent. Pairs &  Tokens (EN) & Sent. Len (EN) & Tokens (ES) & Sent. Len (ES)\\ \hline
Training & 115303 &1365928 &11.8 &1525658 &13.2 \\
Development &14205 &169794 &12.0 &189840 & 13.4 \\
Testing & 14614 & 172850& 11.8&193217 &13.2\\
\hline
\hline 
 $EHR$ & Sent. Pairs & Tokens (EN) & Sent. Len (EN) &  Tokens (ES) & Sent. Len (ES)\\ \hline
Training & 2171 & 34534& 15.9& 36906&17.0 \\
Development &244 & 3946&16.2 & 4191& 17.2 \\
Testing & 595 &10201 &17.1 &10900 &18.3\\
\hline
\end{tabular}
\end{center}
\end{table*}

\paragraph{Implementation Details}
For the implementation we used Pytorch\footnote{https://pytorch.org/}.
All words are kept in the data in our experiments.
We set the number of layers of LSTM to 2 in both the encoder
and the decoder. Other settings are as $hiddensize=128,batchsize=32, dropout=0.2,gradidentclip=5, optimizer=Admda, lr=0.001$.  The hyper-parameter $\alpha$ is tuned from 0 to 1 with interval 0.2, we fixed it as 0.2. During decoding, the beam size is set as 8. All hyper-parameters of our model are tuned using the development set. We use the same data and pre-processing steps for all the models in this work unless noted otherwise.

\paragraph{Baseline models}
We compare our model $NOOV$ to the following two competitive NMT systems: \textbf{Subword NMT} The NMT of rare words with subword units system proposed by \cite{sennrich2015neural} which uses byte pair encoding (BPE) to handle rare words; \textbf{Hybrid NMT} The hybrid word-character NMT system built by \cite{luong2016achieving} that translate mostly at the word level and consult the character components for rare word.

\begin{table}[ht!]
\small
\begin{center}
\caption{\label{Resulst2} BLEU results comparison with baselines on Medline(M) and EHR(EHR) dataset.}
\begin{tabular}{l|c|c|c|c|c|c}
\hline 
\backslashbox{Models}{Tr Data} & \multicolumn{2}{c}{Medline}  & \multicolumn{2}{c}{Medline+EHR} & \multicolumn{2}{c}{ \pbox{5cm}{Medline Pre-train\\ EHR fine-tune}} \\ \hline
   & M&EHR & M&EHR& M&EHR\\\hline 
subword NMT   & 34.50&7.26 & 34.45&30.19& 28.47 &30.03 \\
hybrid NMT &34.70&11.06&34.23&31.36&28.07&31.06\\ \hline
NOOV& \emph{35.18}&6.92 & \textbf{35.90}&33.02&  33.22&\textbf{34.71}\\
\hline
\end{tabular}
\end{center}
\end{table}

As our EHR data is relatively small for NMT systems, we use the $ESPAC_{MedlinePlus}$ dataset to help train the MT models for EHR notes. 
Although both the $ESPAC_{MedlinePlus}$ and the EHR corpora are in the medical domain, they differ significantly. To alleviate the domain divergence, we applied the NOOV and the baseline NMT models to each of the corpora and their joint corpus. For simplicity, we use $Medline$ to refer to the $ESPAC_{MedlinePlus} dataset$. In total the following three experiments data settings are used: \textbf{Medline}: Models are trained on $Medline_{tr}$ and tuned on $Medline_{dev}$; \textbf{Medline+EHR}: Models are trained using $Medline_{tr}$+$EHR_{tr}$ and tuned using $Medline_{dev}$+$EHR_{dev}$; \textbf{Medline Pre-train, EHR fine-tune}: Models are pre-trained on $Medline_{tr,dev}$ and fine-tuned on the $EHR_{tr,dev}$. Results are reported on the test datastes of both $Medline_{te}$ and $EHR_{te}$ for 3 data settings.
\paragraph{Evaluation and results}
We conduct both automatic evaluation (Table 2) and human evaluation.
The human evaluation was done by an English-speaking physician in the US whose first language is Spanish. The physician was provided with the original English sentence, the corresponding gold standard Spanish translation, and three machine translation outputs in a blind fashion and asked to score each output based on two criteria: adequacy and fluency. Adequacy is defined as "how much of the meaning expressed in the gold-standard translation or source is also expressed in the target translation." and scales from 1 to 5 (5 is the best). 
Fluency refers to the target only; the main evaluation criteria are grammar, spelling, choice of words, and style. It also scales from 1 to 5 with 1 as incomprehensible and 5 as flawless. The human evaluation results on \textbf{Adequacy} for the three systems of subword NMT, hybrid NMT and NOOV are respectively 2.78, 2.97 and \textbf{3.26}. Results on \textbf{Fluency} for subword NMT, hybrid NMT and NOOV are respectively 2.64, 2.56, and \textbf{2.84}.

As shown above, the proposed NOOV model beats the two baseline systems in all experiments, judged by both BLEU and by human evaluation. We also conduct error analysis (see Appendix).

\section{Conclusion and future
work}
This paper presents NOOV, a new NMT system that incorporates
a bilingual lexicon and a phrase look-up table  to alleviate the unknown word problem and
also enhance better phrase generation of NMT systems. Our evaluation results show the effectiveness of the NOOV model. In the future we hope to create a large parallel-aligned EHR corpus. We will also explore other models, including domain-adaptation and making further use of the biomedical knowledge resources. 
\bibliographystyle{plain}
\bibliography{emnlp2018}
\newpage
\appendix
\section{Supplemental material}
In this Appendix, we report the error analysis for a better understanding of the proposed system's performance. Our NOOV model addresses the challenges of OOVs in the clinical domain and our results not surprisingly demonstrate the effectiveness of handling unknown words. As shown in example 1 in Table 3, NOOV correctly translated the medical jargon "polyuria" or "dysuria". Even with sentences with broken structures, as shown in Table 3 example 2, despite that de-identification has broken the sentence structure, NOOV has a perfect translation. In contrast, both baseline NMT systems made mistakes.   

NOOV also makes mistakes. Since NOOV is built on the LSTM model, it is not surprising that it failed to translate very long sentences as shown in example 3 in Table 3. To compare NOOV with the baseline NMT models,we further conduct an analysis to see how sentence length impacts the translation results for all three NMT systems. Evaluated on the test set, Figure 1 demonstrates that although when sentences get longer, the BLEU scores of all three systems generally decrease, NOOV outperformed the two baseline NMT systems, especially when the length of the sentences increases. 
Another type of common error we observe is related to grammar. For example, in the case of example 4 in Table 3, NOOV produced good translation of the medical jargon “flu shot” but failed to translate correct articles. 

\begin{table*}[ht!]
\small
\caption{ Two good translated examples by NOOV (Example 1 and 2) and two bad translated examples by NOOV (Example 3 and 4).}
\label{good_example}
\begin{tabular}{l|l}
\hline 

Input & he denies any polyuria or dysuria . \\ 
Ref. & el niega tener poliuria o disuria . \\ 
NOOV & el niega tener poliuria o disuria .\\ \hline

Input & INTERIM history : * * First + name   * * last + name * * returns for followup of his lymphoma .	 \\ 
Ref. & historial INTERINA : * * First + Name   * * Last + Name * * regresa para seguimiento de su linfoma .	\\ 
NOOV & historial interina : * first + name *  * last + name *  regresa para seguimiento de su linfoma .\\ \hline

\hline 

Input & during her hospital stay with the workup for her multiple falls , acute coronary syndrome \\ 
& has been ruled out by EKG   within normal limits and no acute change and troponins have been \\ 
& negative four times and a CVA has been ruled out . \\ 

Ref. & durante su estadía hospitalaria con los estudios para sus caídas múltiples , síndrome  coronario agudo\\ 
 & ha sido descartado  por electrocardiograma { EKG }  dentro de límites normales sin cambio agudo y  \\ 
 & troponinas han sido negativas cuatro veces y un accidente cerebro vascular 
 { CVA } ha sido descartado .\\ 
NOOV & durante su hospitalización durante su período de múltiples hora , se ha descartar el síndrome de células \\
&  coronaria  aguda por dentro de los límites normales y no se ha identificado ningún cambio agudo .\\ \hline

Input & flu shot was provided to patient today . \\ 
Ref. & Vacuna de la Influenza fue suministrada hoy al paciente\\ 
NOOV & vacuna influenza fue suministrado al paciente hoy . \\ \hline

\end{tabular}
\small
\end{table*}

\begin{figure}[ht]
  \centering
  \includegraphics[width=0.50\textwidth]{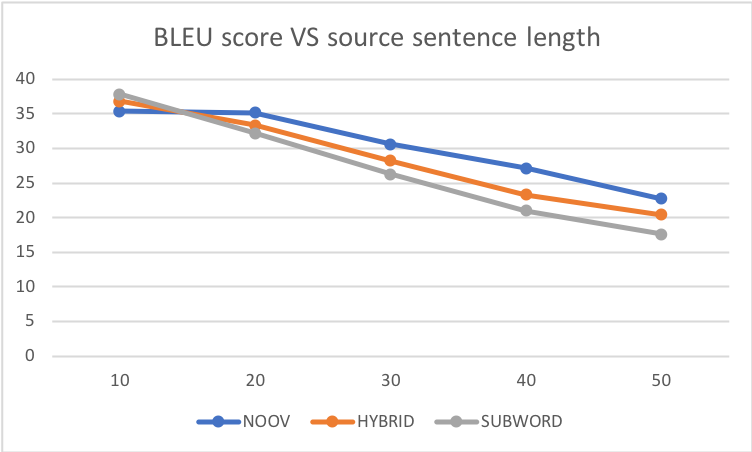}
  \caption{Effects of sentence length on translation results}\label{fig:arch} 
\end{figure}

\end{document}